\documentclass[10pt,twocolumn,letterpaper]{article}

\usepackage[pagenumbers]{cvpr} % To force page numbers, e.g. for an arXiv version

\usepackage[dvipsnames]{xcolor}

\usepackage{algorithm, algorithmic, amsmath, amsfonts, amssymb, mathtools}

\definecolor{cvprblue}{rgb}{0.21,0.49,0.74}
\usepackage[pagebackref,breaklinks,colorlinks,citecolor=cvprblue]{hyperref}

\def\paperID{4328} % *** Enter the Paper ID here
\def\confName{CVPR}
\def\confYear{2024}

\title{CycleGANAS: Differentiable Neural Architecture Search for CycleGAN}

\author{Taegun An\\
Korea University\\
Department of Computer Science\\
{\tt\small antaegun20@korea.ac.kr}
\and
Changhee Joo\\
Korea University\\
Department of Computer Science\\
{\tt\small changhee@korea.ac.kr}
}

\begin{document}
\maketitle
\begin{abstract}

We develop a Neural Architecture Search (NAS) framework for CycleGAN that carries out unpaired image-to-image translation task. Extending previous NAS techniques for Generative Adversarial Networks (GANs) to CycleGAN is not straightforward due to the task difference and greater search space. We design architectures that consist of a stack of simple ResNet-based cells and develop a search method that effectively explore the large search space. We show that our framework, called CycleGANAS, not only effectively discovers high-performance architectures that either match or surpass the performance of the original CycleGAN, but also successfully address the data imbalance by individual architecture search for each translation direction. To our best knowledge, it is the first NAS result for CycleGAN and shed light on NAS for more complex structures.

\end{abstract}    
\section{Introduction}
\label{sec:intro}

Generative Adversarial Networks~\cite{GAN} (GANs) is a unsupervised generative modeling for diverse and realistic data generation with a generator and discriminator trained in an adversarial manner. It can deal with the domain with insufficient amounts of data or high labeling costs and also has a great advantage over the domain where the generation of diverse and high-quality data is critical. In consequence, many variants and extensions of GANs have been proposed for a variety of tasks such as conditional generation, style transfer, machine translation, and anomaly detection.

% Thesis: CycleGAN and its applications + limitations in NAS view
CycleGAN is one of the powerful extensions of GANs and was developed for the image-to-image translation \emph{without image pairing}~\cite{CycleGAN}. Successfully removing the expensive laboring cost of building paired datasets, CycleGAN has been intensively applied to many translation applications, e.g., style transfer, medical diagnosis, voice conversion, etc.
However, CycleGAN requires thorough fine-tuning of multiple neural networks, which has to consider the task objectives and possible imbalances in the dataset. Without the fine-tuning, it is likely to suffer from instability as GANs~\cite{GAN_stability_NIPS2017, GAN_stability_PMLR_mescheder2018}. This often becomes an obstacle when applying CycleGAN to new translation tasks~\cite{Bayesian_CycleGAN} or designing a versatile architecture suitable for multiple tasks.

As a result, for CycleGAN, it is common to have a manual optimization process to tailor its architecture to specific applications~\cite{Assymetric_CycleGAN, CycleGAN_ASR, CycleGAN_VC2}. Such manual optimization is not only very costly but also arbitrarily restricts the scope of the architecture search.

% Thesis: previous NAS for GANs & justification for CycleGANAS
To our best knowledge, there is no priori NAS work for CycleGAN. However, there have been several interesting NAS works for unconditional GANs including AGAN~\cite{AGAN_Wang2019},  AutoGAN~\cite{AutoGAN_Gong_2019_ICCV}, and AdversarialNAS~\cite{Adv_NAS_Gao_2020_CVPR}. AGAN and AutoGAN adopted Reinforcement Learning (RL) to explore architectures through trial-and-error. AdversarialNAS has a more complex cell structure than the others, which motivates it to exploit the gradient-based approach introduced in DARTs~\cite{NAS_DARTS_liu2018} to effectively search on a large search space $(10^{38})$. Although they are an effective NAS method for GANs, their extension to CycleGAN is neither straightforward nor fruitful due to essential differences in the task and structure.
In general, an image-to-image translation task is more complex than an unconditional image generation task, and CycleGAN has twice as many neural networks as GANs. Further, unlike unconditional GANs that can be trained with any random inputs, CycleGAN work only with inputs from finite unpaired datasets and thus should be more sample efficient. The larger search space and the limited inputs are one of the key factors that differentiate the NAS for CycleGAN and the NAS for GANs, and motivate us to develop a novel NAS framework for CycleGAN.

% Thesis: CycleGANAS:  multi-network NAS under data imbalance
% resnet cells + single level / asymetric architecture / performance
In this paper, we propose CycleGANAS, a multi-network architecture search framework for CycleGAN under data imbalance. We design a simple cell inspired by the design of residual cells and build the supernetworks of generators and discriminators of CycleGAN by stacking many simple cells. Accounting for the task nature of CycleGAN and the required search efficiency, we optimized the architecture and neural network weights with CycleGAN objectives \emph{simultaneously}, which is different from the previous NAS for GANs that takes an iterative bi-level optimization method. Finally, during the search, we let the two generators have a different architecture, allowing them to be better tailored to each subtask and dataset.
Through the experiments on various unpaired datasets, we show that CycleGANAS searches for good architectures not only efficiently but also in a stable manner, even under the data imbalance.

Our contributions can be summarized as follows.
\begin{itemize}
    % cont. 1. framework
    \item We develop CycleGANAS, a novel framework of multi-network architecture search for CycleGAN under data imbalance. We stack up many simple ResNet-based cells and take the gradient-based approach along with single-level joint optimization of neural network architecture and weights. Our framework admits asymmetric architectures that take into account the data imbalance.
    % cont. 2. experiments & performances
    \item We show the performance of CycleGANAS through extensive experiments with various unpaired datasets. We investigate the effect of CycleGAN architecture and model size, demonstrating that manual balancing or naive asymmetric models are not effective for data imbalance. In contrast, we observe that CycleGANAS successfully searches for good architectures in a stable manner and achieve high performance under the data imbalance.
\end{itemize}

\section{Related Works}
\label{sec:Related Works}

\subsection{Neural Architecture Search (NAS)}
% introduce NAS
Neural architecture search (NAS) is a big branch of automated machine learning (Auto-ML) to search for the best neural network design, taking into consideration the task and dataset. NAS studies are commonly classified using three criteria: the search space, search method, and evaluation method~\cite{NAS_survey_He_2021, NAS_Survey_JMLR}. Among these, the search method is the key element for identifying various NAS algorithms. In the following, we provide a concise overview of NAS algorithms in the literature from the perspective of the search method.

% Thesis: NAS methods -- updated papers :)
Several search methods have been developed for NAS including RL, evolutionary algorithms (EA), differentiable methods, and other optimization techniques. Since RL was successfully adopted for NAS in~\cite{NAS_RL_zoph2017}, it has been exploited to optimize many architectures including CNNs~\cite{NAS_RL_baker2017, NAS_RL_cai2018, NAS_RL_ENAS}. Although they outperform the handcrafted neural networks in performance, they demand a significant amount of computation.
Although recent NAS frameworks with Bayesian optimization (BO)~\cite{NAS_BO_kandasamy2018, BANANAS_2021, BayesNAS_pmlr-v97-zhou19e, NAS_BO_ru2021, ru2021interpretable} or evolutionary algorithms (EA)~\cite{NAS_EA_AJ2019, NAS_EA_peng2020, NAS_EA_Liang2021} substantially reduce the computation cost, their applications are still limited to a problem with small search space. Gradient-based NAS method, first appeared in DARTS~\cite{NAS_DARTS_liu2018}, enables NAS with a large search space through continuous relaxation of architectures. It has been reported that in certain problems, the architecture converges to the optimal one under the gradient-based NAS~\cite{NAS_CG_li2021}.

% Thesis: usage of GAN framework / GAN for image-to-image translation: architectural extension / emphasis on our work
\subsection{Generative Adversarial Networks (GANs)}
Since the GANs framework has been developed for the unconditional image generation via adversarial training of generator and discriminator, the framework has been extended to many computer vision tasks including super-resolution~\cite{SRGAN, ESRGAN}, image inpainting~\cite{GAN_inpainting}, natural language processing (NLP)~\cite{GAN_Voice_2017, GAN_Voice_ICASSP}, etc.

CycleGAN is also an extension of the GANs framework to the multi-network system for image-to-image translation task. Pix2pix~\cite{pix2pix} is the first work that performs the translation task with two pairs of GANs and paired data. CycleGAN~\cite{CycleGAN} and DiscoGAN~\cite{DiscoGAN} remove the requirement of the dataset pairing by introducing the cycle-consistency objective, and enable the translation with \emph{unpaired data}.
The technique has been now widely used for medical image translation~\cite{CycleGAN_medical}, frame prediction~\cite{CycleGAN_prediction}, inter-domain translation~\cite{CycleGAN_interdomain}, and multi-modal learning~\cite{BiCycleGAN_NIPS2017}.

%\RED{some recent manual CycleGANs?}

% Thesis: NAS for GANs - RL methods / limitations
\subsection{NAS for GANs}
Since the success of NAS for convolutional neural networks (CNNs), NAS for GANs has attracted much attention. As in CNNs, RL-based search methods were adopted in NAS for GANs~\cite{AGAN_Wang2019, AutoGAN_Gong_2019_ICCV, E2GAN_Tian_2020_ECCV}, in which the architecture of GANs is divided into a predetermined number of cells and optimized through an LSTM agent. A common challenge of these RL-based approaches is the computational complexity to obtain Inception Score (IS) or Frechet inception distance (FID) score, which is used as the reward feedback. Some works introduce score predictors to mitigate the computation burden~\cite{Yi_2023_CVPR}.

% Thesis: AdversarialNAS & limitations - just state the differences, and show it in 3.2 one-step NAS formulation
The most relevant to our work among previous NAS frameworks for GANs is AdversarialNAS~\cite{Adv_NAS_Gao_2020_CVPR} that adopts the gradient-based, differentiable architecture search method.
%\TG{gradient-based optimization as search method (or) differentiable architecture search method}. 
Using the adversarial loss, AdversarialNAS could achieve state-of-the-art performance on a large search space at the cost of $1$~GPU day. However, the extension of AdversarialNAS to CycleGAN is not straightforward, since it has a large and complicated cell structure, which makes it hard to scale, and its bi-level optimization is not sufficiently efficient for the task with a small amount of data, i.e., the image-to-image translation task of CycleGAN~\cite{Saxena_2023_CVPR}.

% Thesis: NAS(?) and CycleGAN
While not directly related to NAS, there are a few studies that focus on compressing the architecture of CycleGAN via combinatorial optimization~\cite{CycleGAN_CO_Compression} or evolutionary algorithm~\cite{CycleGAN_EA_Compresssion}. Given good reference architectures, they find small-size architectures with comparable performance. Although they are doing a sort of architecture search, their approach is quite different from NAS since they have clear reference models.
\section{CycleGANAS -- NAS for CycleGAN}

\subsection{Preliminaries}

\begin{figure}[t]
    \begin{center}
    \centerline{\includegraphics[width=1.0\linewidth]{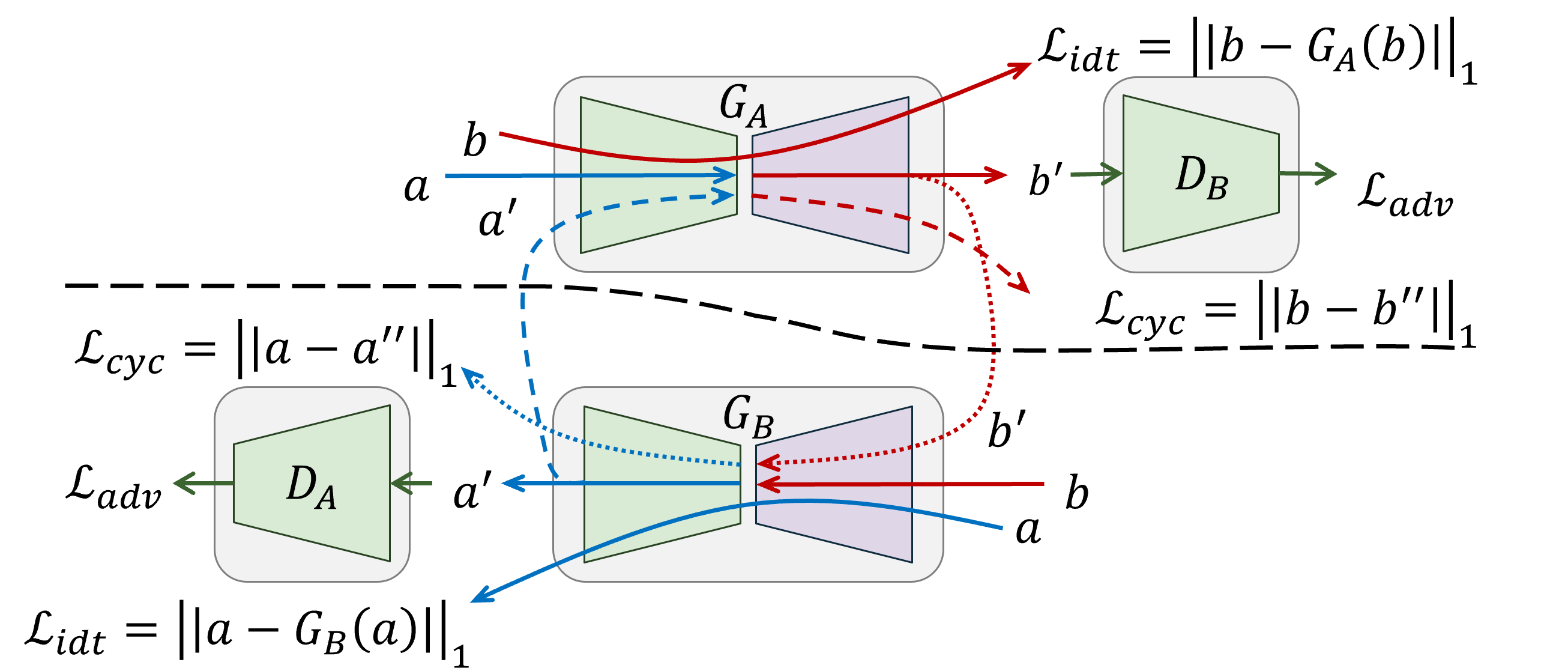}}
    \caption{Overall process of CycleGAN and its losses. CycleGAN uses two generators $G_A, G_B$ to translate image $a\in \mathcal{A}$ to image $b' \in \mathcal{B}$ and image $b \in \mathcal{B}$ to image $a' \in \mathcal{A}$, and two discriminators $D_A, D_B$ to distinguish the generated images. It computes several losses using the images from different sources.
    }
    \label{fig:CycleGAN}
    \vspace{-0.5cm}
    \end{center}
\end{figure}

% Thesis: explain CycleGAN
CycleGAN has two generators ($G_A, G_B$) and two discriminators ($D_A, D_B$) as shown in Fig.~\ref{fig:CycleGAN}, and translate images in dataset, which consists of two subdatasets, each from a different domain denoted by $\mathcal{A}, \mathcal{B}$. We slightly abuse the notation and also denote the subdataset of each domain by $\mathcal{A}$ and $\mathcal{B}$, respectively. CycleGAN aims to learn the mapping of optimal generators $G^*_{A}:\mathcal{A} \rightarrow \mathcal{B}$, $G^*_{B}:\mathcal{B} \rightarrow \mathcal{A}$, and optimal discriminators $D^*_A: \Omega \rightarrow \{0,1\}, D^*_B: \Omega \rightarrow \{0,1\}$ that distinguish between translated samples and real samples, where $\Omega$ denotes the set of all possible image samples. For CycleGAN, we relax the output of discriminators to a real number in the range $[0,1]$, and consider $G_A, G_B, D_A, D_B$ as a function that takes an input image and outputs an image or a number in $[0,1]$. The objective of CycleGAN consists of three loss functions, known as the adversarial, cycle-consistency, and identity loss~\cite{CycleGAN}.

\begin{itemize}
\item The adversarial objective makes each pair of generator and discriminator engage in a two-player mini-max game. The \emph{adversarial loss} $L_{adv}(G_A, D_B)$ for $G_A, D_B$ can be written as
\begin{equation}\label{eqn:CycleGAN_adv_loss}
    \begin{split}
        L_{adv}(G_A, D_B) &= \mathbb{E}_{b \sim \mathcal{B}}[\log D_B(b)] \\
                          &+ \mathbb{E}_{a \sim \mathcal{A}}[\log(1 - D_B(G_A(a)))].
    \end{split}
\end{equation}
\noindent The adversarial loss $L_{adv}(G_B, D_A)$ for $G_B$ and $D_A$ can be defined similarly. 

% explanation: Cycle-consistency
\item It is claimed that CycleGAN should be cycle-consistent, i.e., $G_B(G_A(a)) \approx a$ for $a \in \mathcal{A}$ and $G_A(G_B(b)) \approx b$ for $b \in \mathcal{B}$. The cycle consistency objective is imposed only on the generators and couples them under the cooperative framework to generate better output images for each other. The \emph{cycle-consistency loss} $L_{cyc}(G_A, G_B)$ can be written as

\begin{equation}\label{eqn:CycleGAN_cyc_loss}
\begin{split}
    L_{cyc}(G_A, G_B) &= \mathbb{E}_{a \sim \mathcal{A}}[|G_B(G_A(a)) - a|] \\
                      &+ \mathbb{E}_{b \sim \mathcal{B}}[|G_A(G_B(b)) - b|]. %
\end{split}
\end{equation}

% explanation: Identity
\item It has been shown that, if the input image does not belong to the target domain, the identity objective helps the generator to preserve the identity of an input image, e.g., color composition. The \emph{identity loss} $L_{idt}(G_A)$ for $G_A$ can be written as

\begin{equation}\label{eqn:CycleGAN_idt_loss}
    L_{idt}(G_A) = \mathbb{E}_{b \sim \mathcal{B}}[|G_A(b) - b|].
\end{equation}

The identity loss for $G_B$ can be written similarly.
\end{itemize}

% explanation: final objectives
The full objective of CycleGAN is formed by linearly combining the three loss functions as
\begin{equation}\label{eqn:loss_comb}
    \begin{split}
        &L(G_A, G_B, D_A, D_B) \\
        &= \lambda_1 L_{adv}(G_A, D_B) + \lambda_2 L_{adv}(G_B, D_A) \\
        &+ \lambda_3 L_{cyc}(G_A, G_B) + \lambda_4 L_{idt}(G_A) + \lambda_5 L_{idt}(G_B),
    \end{split}
\end{equation}
where $\lambda_1,\dots,\lambda_5$ are a weight. Through this work, we set $\lambda = (\lambda_1,\dots,\lambda_5)=(1,1,10,5,5)$ unless otherwise stated. Note that the discriminators are involved only in the adversarial losses.

% Thesis: 1st contribution: resnet-based cell design and supernetwork
\subsection{Search space}\label{sec:searchspace}

%Based on the overall process of CycleGAN, 
We develop CycleGANAS that simultaneously searches neural network architectures for $G_A, G_B, D_A, D_B$. Considering the significant challenge in conducting architecture search from scratch, many NAS approaches restrict their search space to a combination of operations referred to as a \emph{cell}, and construct a neural network architecture by stacking a few cells~\cite{AutoGAN_Gong_2019_ICCV, Adv_NAS_Gao_2020_CVPR}. We also follow the cell-based NAS approach, but different from the previous works, we use a much simpler cell structure and stack many of them to construct an architecture.

Motivated by the neural network architectures of CycleGAN, we design a ResNet-based cell for CycleGANAS. Our cell has only two operations, each of which followed by a normalization layer and an activation layer.
We restrict the cell's operations by \emph{cell type}, which can be either encoding $e$, residual $r$, or decoding $d$. Letting $\mathcal{S}^T$ be the set of possible operations for the cell type $T \in \{e,r,d\}$, we define the operation sets as
\begin{itemize}
    \item $\mathcal{S}^e = \{$max pooling, avg pooling, Conv3x3, Conv4x4, Conv5x5, Conv7x7, DilConv3x3, DilConv5x5$\}$,
    \item $\mathcal{S}^r = \{$Conv3x3, Conv5x5, Conv7x7, DilConv3x3, DilConv5x5$\}$,
    \item $\mathcal{S}^d = \{$Nearest neighbor interpolation, Bi-linear interpolation, Transposed Conv3x3$\}$.
\end{itemize}
A cell $C^T$ of type $T$ has two operations from $\mathcal{S}^T$.

We build a generator $G$ of $N$ cells; starting from one encoding cell, followed by $N-2$ residual cells, and ending with one decoding cell. Thus, we can represent a generator as a sequence of cells: $G = (C^e_1, C^r_2, \dots, C^r_{N-1}, C^d_N)$. For a decoder $D$, we have it with a much simpler structure of $2$ encoding cells: $D = (C^e_1, C^e_2)$.
Note that a generator has the search space of size $|\mathcal{S}^e|^2 \cdot |\mathcal{S}^r|^{2(N-2)} \cdot |\mathcal{S}^d|^2$, where $|\cdot|$ denotes the cardinality. We will use $N=11$ as a default setting for each generator, resulting in $8^{2} \times 5^{18} \times 3^{2}\approx 2.2\times 10^{15}$. Similarly, each discriminator has the search space of size $8^{4}$. Since there are two generators and two discriminators, the total search space size of CycleGANAS is $8.1 \times 10^{37}$, which is comparably large considering those of AutoGAN ($10^{5}$) and AdversarialNAS ($10^{38}$).

\begin{figure}[t]
    \begin{center}
    \centerline{\includegraphics[width=1.0\linewidth]{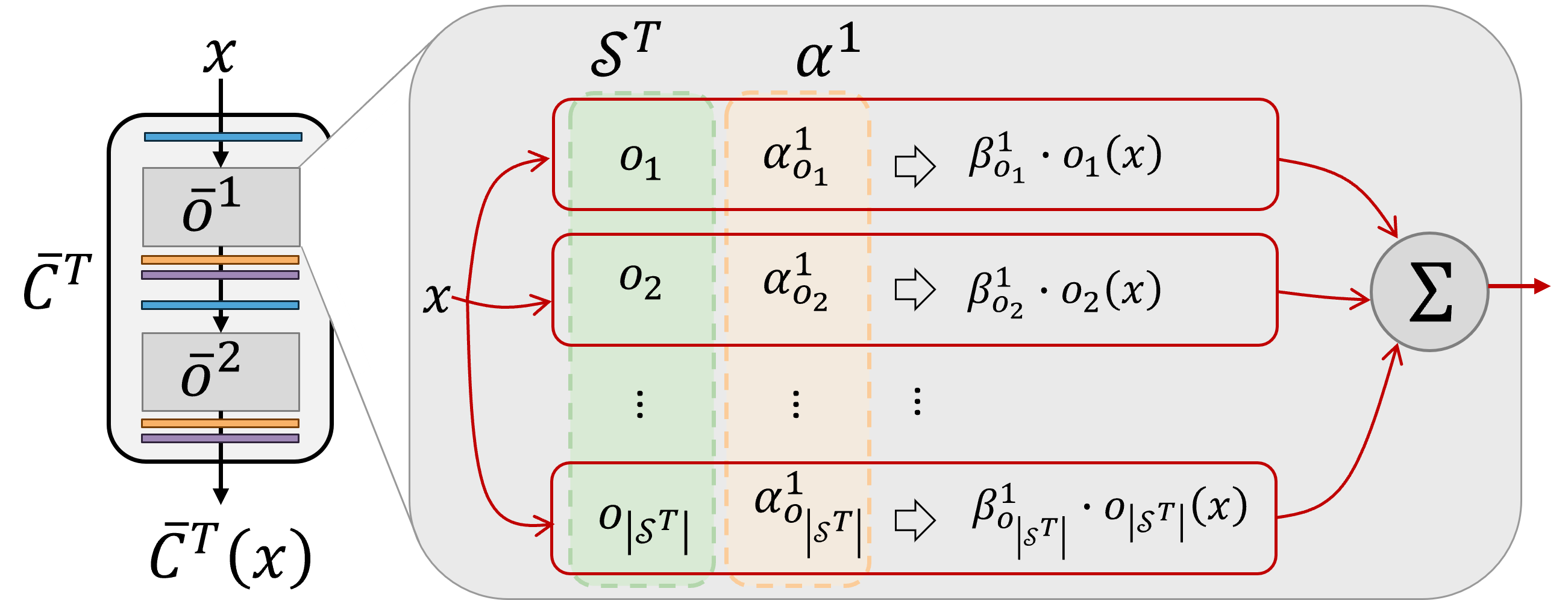}} % add some sub-operation boxes?
    \caption{A super-network cell with two mixed operations $\bar{o}^1, \bar{o}^2$. The weights $\alpha$ of the mixed operations are trained by the gradient descent. After the architecture search, each cell is converted to an ordinary cell with two discrete operations.}
    \label{fig:cell}
    \vspace{-0.5cm}
    \end{center}
\end{figure}

Element-wise searching in the huge space will demand prohibitively large computational cost. To avoid excessive computation, we adopt the idea of \emph{mixed operation} from DARTS~\cite{NAS_DARTS_liu2018} and take the approach of \emph{differentiable neural architecture search}. Basically, we convert the discrete search space into a continuous one by replacing an operation with a mixed one of multiple operations that admits the gradient-based search. 
To elaborate, consider a cell $\bar{C}^T$ that has two mixed operations $\bar{o}^1, \bar{o}^2$ in order, where each $\bar{o}^i$ with $i \in {1,2}$ is a combination of all possible operations, i.e., 
\[
    \bar{o}^{i} =  \sum_{o \in\mathcal{S}^T } \beta^i_o \cdot o, 
\]
where $\beta^i_o = \frac{\exp(\alpha^{i}_o)}{\sum_{p \in \mathcal{S}^T}\exp(\alpha^{i}_{p})}$ and 
$\alpha^i_o$ is the weight of operation $o \in \mathcal{S}^T$, as shown in Fig.~\ref{fig:cell}.
Also, suppose that we build a super-network by stacking the cells with mixed operations. 
Then we can optimize all $\alpha^i$'s of the super-network's cells through gradient descent. When the search finishes, we construct a discrete architecture by selecting the highest-weight operations in the super-network, i.e., each cell $\bar{C}^T = \{\bar{o}^1, \bar{o}^2\}$ of the super-network is converted to an ordinary cell $C^T = \{o^1,o^2\}$, where $o^i = \arg\max_{o \in \mathcal{S}^T} \alpha^{i}_{o}$.

Architecture search with the super-network demands a substantial amount of memory and time to take into account all the possible operations, which can be burdensome in practice. To this end, it is common that, for the search, one uses the super-network with reduced hidden dimension, and after the search, builds the discrete architecture with restored hidden dimension~\cite{Adv_NAS_Gao_2020_CVPR}.
We also apply the technique of the \emph{hidden dimension reduction} to CycleGANAS for the architecture search. For example, during the search, a (super-network) generator of our CycleGANAS takes an image input of $256\times 256\times 3$, and encodes it to a tensor of $64 \times 64 \times 64$, whose hidden dimension size is smaller than that of the original CycleGAN's encoder output ($64 \times 64 \times 256$). Passing through the residual blocks, the generator decodes it back to the shape of $256\times 256\times 3$. Once the search completes, from the trained super-network, we construct a discrete architecture, whose encoder output has the shape of $64 \times 64 \times H$, where the hidden dimension $H$ is a hyperparameter.

\subsection{Optimization process}
%\RED{support more on single-level optimization}

\begin{algorithm}[tb]
\caption{One-step CycleGANAS.}
\label{alg:CycleGANAS}
\textbf{Initialize weights}: $\theta_{G_A}, \theta_{G_B}, \theta_{D_A}, \theta_{D_B}$\\
\textbf{Initialize differentiable architecture}: $G, G, D, D$\\ % notation issue...?
\textbf{Input}: Unpaired dataset $\mathcal{A, B}$\\
\textbf{Output}: architecture of $G_A,G_B,D_A,D_B$ % why this make center-align? - below empty line solves the problem

\begin{algorithmic}[1] %[1] enables line numbers
\FOR{each search epoch}
    \FOR{each iteration}
        \STATE sample $a\in\mathcal{A}$ and $b\in\mathcal{B}$ \\
        \smallskip
        \STATE /* Forward path */
        \STATE Compute $L_{adv}(G_A, D_B), L_{adv}(G_B, D_A)$
        \STATE Compute $L_{cyc}(G_A, G_B), L_{cyc}(G_B, G_A)$
        \STATE Compute $L_{idt}(G_A), L_{idt}(G_B)$
        \STATE Compute $L(G_A, G_B, D_A, D_B)$ \\
        
        \smallskip
        \STATE /* Backward path */
        \STATE Update $\theta_{G_A}, \theta_{G_B}$ from $L(\cdot)$ 
        \STATE Update $\theta_{D_A}, \theta_{D_B}$ from corresponding $L_{adv}(\cdot)$
    \ENDFOR
\ENDFOR
\STATE \textbf{return} architecture of $G_A,G_B,D_A,D_B$ % =0 ?????
\end{algorithmic}
\end{algorithm}

% Thesis: notations and bi-level optimization
Let $\alpha$ denote the vector of all architecture weights, and let $w$ denote the vector of neural network weights. Note that if $\alpha$ changes, the optimal $w$ also changes, and vice versa.
Typically, a NAS for GANs takes a bi-level optimization process to optimize $\alpha$ and $w$, which is equivalent to iteratively apply the following two equations in sequence,
\begin{equation}\label{eqn:bilevel_optimization}
\begin{split}
    \alpha^{*} 
        &= \textstyle \arg\min_{\alpha} L_{val}(\alpha, w^{*}), \\
    w^{*}      
        &= \textstyle \arg\min_{w}L_{train}(\alpha^*,w), 
\end{split}
\end{equation}
\noindent where $L_{val}$ is a loss with the validation dataset and $L_{train}$ is a loss with the training dataset. For example, AutoGAN has the FID score for $L_{val}$ and the adversarial loss for $L_{train}$, and AdversarialNAS has the adversarial loss for both $L_{val}$ and $L_{train}$.

We highlight that the bi-level optimization~\eqref{eqn:bilevel_optimization} requires the computation of $L_{train}$ and $L_{val}$ from two exclusive datasets for stability. AutoGAN and AdversarialNAS can easily accomplish this since the input to the GANs generator is a random sample from Gaussian noise.
For CycleGAN, however, the input to a generator is not a random sample but an image from a subdataset, and thus each input subdataset has to be divided into two for the bi-level optimization. Further, rotating the role of divided subdatasets will be necessary; otherwise, the architecture optimizer and weight optimizer will only observe a portion of the entire subdataset, leading to a performance degradation. As a result, the bi-level optimization process of CycleGANAS requires a more intricate approach to dividing the dataset and rotating their roles.

We simplify the optimization process by developing a single-level optimization process of $\alpha$ and $w$. We use the entire input subdataset for the joint optimization of
\begin{equation}\label{eqn:single_level_optimization}
    \textstyle (\alpha^*,w^*) = \arg \min_{\alpha,w} ~L_{train}(\alpha,w),
\end{equation}
where $L_{train} = L(G_A,G_B,D_A,D_B)$ is the CycleGAN loss~\eqref{eqn:loss_comb}.
Note that the single-level optimization not only eliminates the need for dividing the input subdataset but also reduces the number of optimization steps by half, in comparison to the bi-level optimization. This results in an accelerated learning process.
We denote CycleGANAS with the single-level optimization by \emph{one-step} CycleGANAS, whose detailed process is shown in Algorithm~\ref{alg:CycleGANAS}, where $\theta = (\alpha, w)$.

\section{Experiments}
% Explain the datasets: basic information, horse2zebra imbalances and more.
We evaluate CycleGANAS with several unpaired datasets, e.g. maps, facades, apple2orange, horse2zebra, summer2winter, and iphone2dslr-flower.%, as in~\cite{CycleGAN}. 
Each dataset has two subdatasets, each of which may have a different number of images, and all the images are of the same shape $3\times256\times256$. For certain datasets, there exists an imbalance in the data, and the image translations between the two subdatasets have different levels of difficulty. For example, in the horse2zebra dataset, the number of pixels taken by zebra is more than twice as many as those taken by horse~\cite{HZ_imbalance}. This data imbalance demands asymmetric capability of neural networks, in particular, of the two generators. In general, a more challenging translation task, e.g., zebra-to-horse, is likely to necessitate a larger generator model\footnote{We remark that the original CycleGAN is symmetric -- its two generators have the same architecture, and the same for the two discriminators.}. We demonstrate that CycleGANAS not only successfully search high-performance architectures, but also, in the presence of dataset imbalance, naturally adopts asymmetric architectures, effectively mitigating the issue.

% Explain settings: hyperparameter, hardware, software
Throughout our experiments, we mostly use the configuration of CycleGAN~\cite{CycleGAN}, e.g., batch size $1$ and instance normalization. For one-step CycleGANAS of Algorithm~\ref{alg:CycleGANAS}, we use Adam optimizer with learning rate of $\alpha = 0.0002$, $\beta_{1} = 0.5$, $\beta_{2} = 0.999$, and set the maximum search epoch to $400$. All our experiments are based on Python 3.8, CUDA 11.3, CuDNN 8.2.0, and the learning frameworks are implemented with PyTorch. To evaluate architectures, we use the FID score estimation provided by the clean-fid project~\cite{clean_fid_parmar2021}. Since the FID score is stochastic for an architecture~\cite{GAN_stochastic_FIDs}, we repeat the weight training several times and select the one with the best (lowest) FID score.

% Explain the order of experiments
We present the impact of the optimization method on performance, and evaluate the performance of CycleGANAS using diverse datasets and illustrate its response to data imbalance. 

\subsection{Bi-level vs single-level optimization}

\begin{figure}[t]
    \begin{center}
    \centerline{\includegraphics[width=1.0\linewidth]{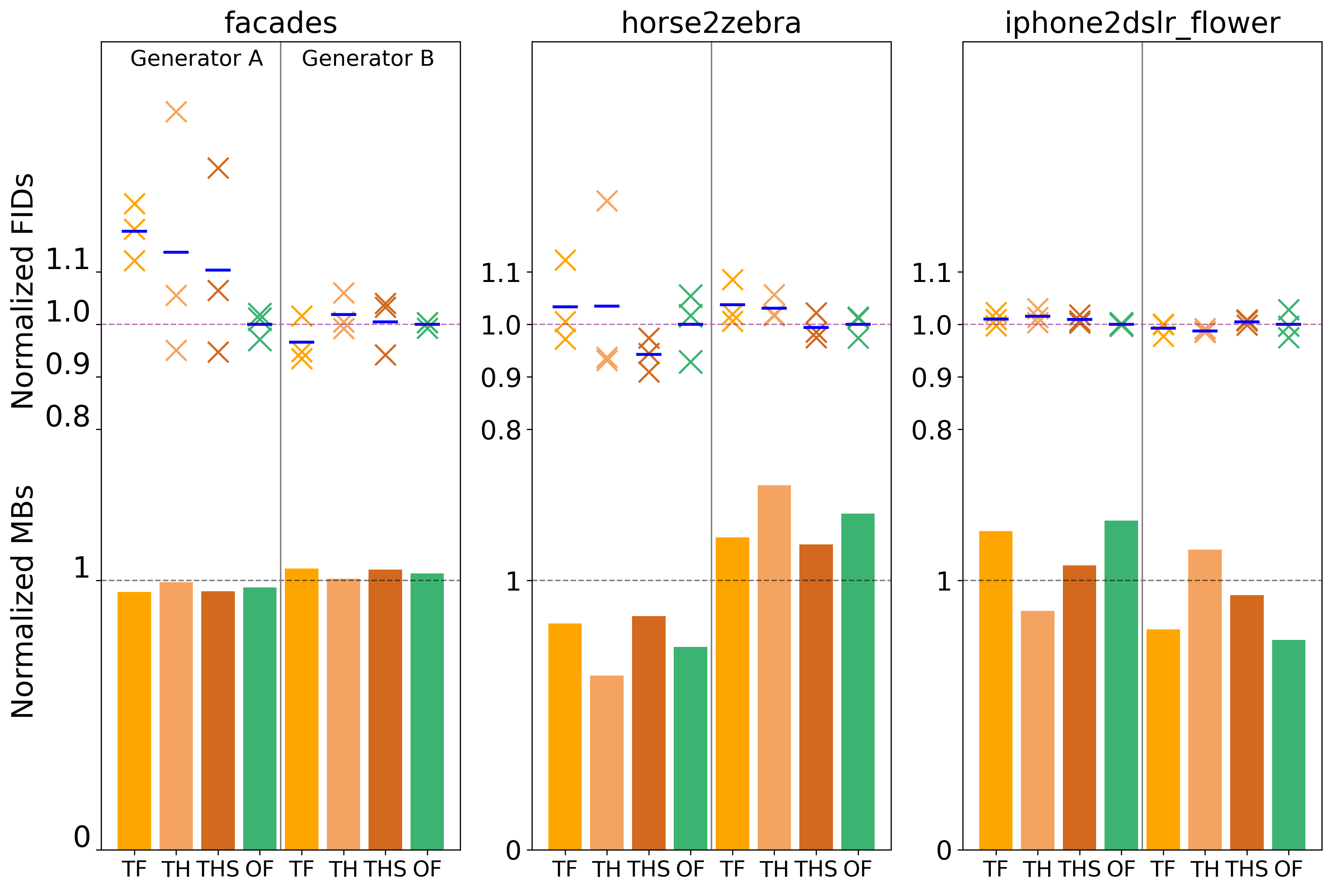}}
    \caption{Performance of the four variants of CycleGANAS with three datasets, in terms of the generator model size and FID scores. The model sizes are normalized w.r.t. that of the original CycleGAN's generator, and the FIDs w.r.t. the average FIDs of one-step CycleGANAS (OF). For each architecture outcome, we repeat the weight training $3$ times, and mark the FIDs by cross and the average by blue dash.
    %\CJ{Also add 0.8, 0.9, ... for FIDs. + How about normalize the FIDs w.r.t average FIDs of OF (why this one only by `min'?)}\TG{was not sure between mean vs min... modified figure!}
    }
    \label{fig:BvsS}
    \vspace{-0.5cm}
    \end{center}
\end{figure}

%%%%%%%%%%%%%%%%%%%%%%%%%%%%%%%%%%%
\begin{figure*}[t]
    \begin{center}
    \centerline{\includegraphics[width=0.95\linewidth]{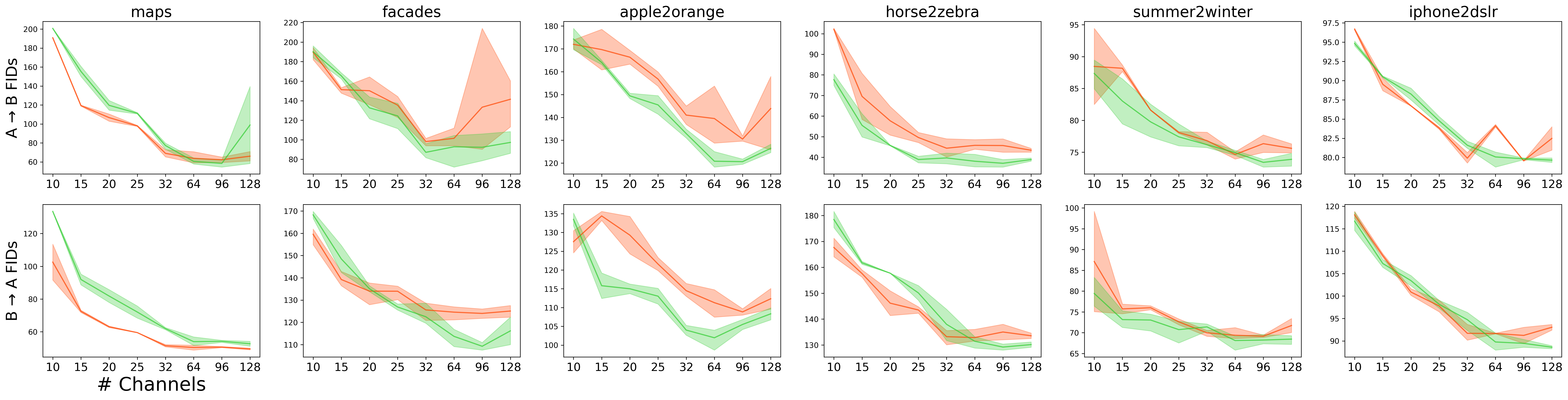}} % add some sub-operation boxes?
    \caption{Performance of CycleGAN (red) and the architecture outcomes of CycleGANAS (green) with different hidden dimension sizes. We evaluate each architecture $3$ times. The mean (thick line) and range (light shade) of FIDs are shown. The outcomes of CycleGANAS achieve comparable performance or even outperforms CycleGAN.}
    \label{fig:main1}
    \vspace{-0.5cm}
    \end{center}
\end{figure*}
%%%%%%%%%%%%%%%%%%%%%%%%%%%%%%%%%%%

\begin{figure*}[t]
    \begin{center}
    \centerline{\includegraphics[width=0.95\linewidth]{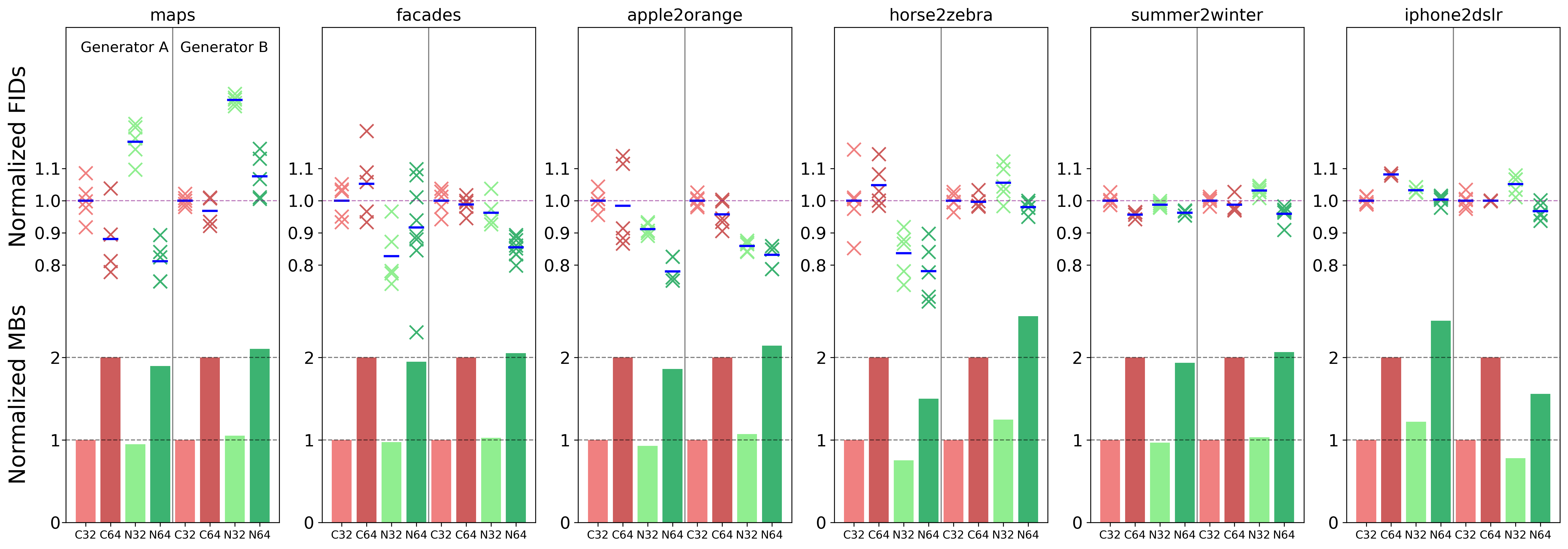}} % add some sub-operation boxes?
    \caption{Performance of CycleGAN (red) and the architecture outcomes of CycleGANAS (green). Two CycleGAN architectures with $H=32$ ($C32$) and $H=64$ ($C64$) are used. We also use two versions of CycleGANAS architecture outcomes ($N32$ and $ N64$), such that their total model sizes are roughly the same as $C32$ and $C64$, respectively. Comparing their performance over 6 datasets, we can observe that the architectures searched by CycleGANAS outperform CycleGAN, in particular, for the larger model size.}
    \label{fig:size_ratio}
    \vspace{-0.5cm}
    \end{center}
\end{figure*}

Besides the single-level optimization, we can incorporate our CycleGANAS framework with the previous bi-level optimization, which is denoted by \emph{two-step} CycleGANAS. Numerous variations of the two-step CycleGANAS can emerge based on how the subdataset is divided and rotated. We consider the following three variants of two-step CycleGANAS.

\begin{itemize}
    \item (TF) Two-step CycleGANAS with the full subdataset: Without dividing the subdatasets, we use the bi-level optimization of $w$ using $\mathcal{A}$ and $\alpha$ using $\mathcal{B}$.
    \item (TH) Two-step CycleGANAS with evenly halved subdatasets: For $\mathcal{A}$, we assign all its images to $\mathcal{A}_{1}$ or $\mathcal{A}_2$ at random  such that $|\mathcal{A}_1| = |\mathcal{A}_2|$. Similarly, we have $\mathcal{B} \rightarrow \mathcal{B}_{1}, \mathcal{B}_{2}$. Then, we use the bi-level optimization of $w$ using $\mathcal{A}_1,\mathcal{B}_1$ and $\alpha$ using $\mathcal{A}_2,\mathcal{B}_2$ throughout the search.
    \item (THS) Two-step CycleGANAS with evenly halved subdatasets and swapping: It is similar to Two-half, except that after certain epoch, we swap the role of the halved subdatasets. Our intention is for the optimizers of $w$ and $\alpha$ to have visibility over all images within the entire subdatasets. In our experiments, we do the bi-level optimization of $w$ using $\mathcal{A}_1,\mathcal{B}_{1}$ and $\alpha$ using $\mathcal{A}_2,\mathcal{B}_{2}$ for the first $200$ epochs, and then do the optimization of $w$ using $\mathcal{A}_2, \mathcal{B}_{2}$ and $\alpha$ using $\mathcal{A}_1, \mathcal{B}_{1}$ afterward.
\end{itemize}
Unlike the two-step CycleGANAS, our one-step CycleGANAS features a simpler optimization process and necessitates only half the iterations.
\begin{itemize}
    \item (OF) One-step CycleGANAS with the full subdatasets. We jointly optimize $w$ and $\alpha$ using $\mathcal{A}, \mathcal{B}$.
\end{itemize}

We search CycleGAN architectures using the above four schemes with three different datasets of facades, horse2zebra, and iphone2dslr. Fig.~\ref{fig:BvsS} shows the model size of the generators and FIDs of their searched architectures, which are normalized by the model size of the original CycleGAN's generator, and by the minimum FIDs of one-step OF CycleGANAS, respectively.

% architecture: data imbalance
In the facades dataset, the model sizes are all similar and the best (lowest) FID is achieved by two-step THS. On the other hand, one-step OF achieves the least variation of FIDs and the lowest average, which implies that one-step OF performs well in a stable manner. In horse2zebra with data imbalance, all schemes effectively discover asymmetric architectures, with a focus on enhancing the generator for zebra-to-horse translation, although the extent of asymmetry varies among the schemes. Also we can observe that two-step THS and one-step OF outperform two-step TH. We conjecture that it is due to the limited visibility of two-step TH's optimizers to the subdatasets.
In iphone2dslr, CycleGAN is asked to increase or decrease the image resolution, and it is clear that the resolution-increasing task is more difficult than the other. Similar to the horse2bebra case, most schemes except two-step TH provide asymmetric architectures enhancing the generator for iphone-to-dslr translation. Further, one-step OF achieves the best FID scores. Another interesting observation is that the reversely asymmetric architectures searched by two-step TH also achieve comparable performance.

Overall, two-step TF/TH/THS schemes exhibit a high variation depending on the translation task, while one-step OF achieves a competitive, close-to-best result in a stable manner. Further it is worth highlighting that one-step OF completes the optimization process in half the iterations of two-step methods. Given its performance and simplicity, we establish it as the default choice for CycleGANAS. Henceforth, when referring to CycleGANAS, we are referring to one-step CycleGANAS.

\subsection{Performance evaluation}

We demonstrate the search capability of CycleGANAS. Over $6$ datasets, we evaluate the architectures searched by CycleGANAS in comparison with the original CycleGAN. We vary the hidden dimension size $H$ for both CycleGAN and the search outcomes of CycleGANAS to see their achievable performance. For each architecture, we repeat its evaluation $3$ times.

Fig.~\ref{fig:main1} shows the mean (thick line) and range (light shade) of the FID scores of CycleGAN (red) and the architecture outcomes of CycleGANAS (green). Overall, in terms of the lowest FIDs, the outcomes of CycleGANAS demonstrate comparable performance (for maps and iphone2dslr) or even outperform CycleGAN (for facades, apple2orange, horse2zebra, and summer2winter). Another interesting observation is that a larger $H$ does not imply a better performance, and there is an optimal value, usually between $32$ and $96$, depending on the dataset and architecture.

\noindent Remark: the architecture of CycleGAN and the outcomes of CycleGANAS have convolutions of different filter sizes, leading to differences in their model size in bytes even when they have the same hidden dimension size $H$.

Next we conduct a more direct performance comparison between CycleGANAS outcomes and CycleGAN by ensuring their model sizes are equal. From the previous experiment results, we train two CycleGAN architectures with $H=32$ and $64$, denoted by $C32$ and $C64$, respectively. For the architectures searched by CycleGANAS, we adjust the hidden dimension sizes accordingly such that the total model size roughly equals that of $C32$ or $C64$. Note that we configure the two generators searched by CycleGANAS to have the same hidden dimension size $H$, and as a result, depending on their chosen convolution operations, they will have a different model size, leading to asymmetric architectures.

Fig.~\ref{fig:size_ratio} shows the experiment results over $6$ datasets in terms of generator model sizes and FIDs. All the model sizes are normalized with respect to that of $C32$ generator ($11.378$ MB), and the FIDs with respect to $C32$'s mean FIDs. $N32$ and $N64$ denote the architectures searched by CycleGANAS with normalized $H$ with respect to $C32$ and $C64$, respectively. Overall, CycleGANAS successfully finds good architectures for most datasets (except maps). Its architectures have comparable (for summer2winter and iphone2dslr) or better performance (for facades, apple2orange, horse2zebra) than the CycleGAN counterpart. In particular, with the larger model size, the outcomes of CycleGANAS outperform the original CycleGAN, by up to $30$\% in $\mathcal{A}$-to-$\mathcal{B}$ and $10\%$ in $\mathcal{B}$-to-$\mathcal{A}$. Further we can also observe that, for the datasets with imbalance (horse2zebra and iphone2dslr), CycleGANAS provides asymmetric generator architectures accordingly, and for the others, it yields the generators of similar model size.

\begin{figure}[t]
    \begin{center}
    \centerline{\includegraphics[width=0.9\linewidth]{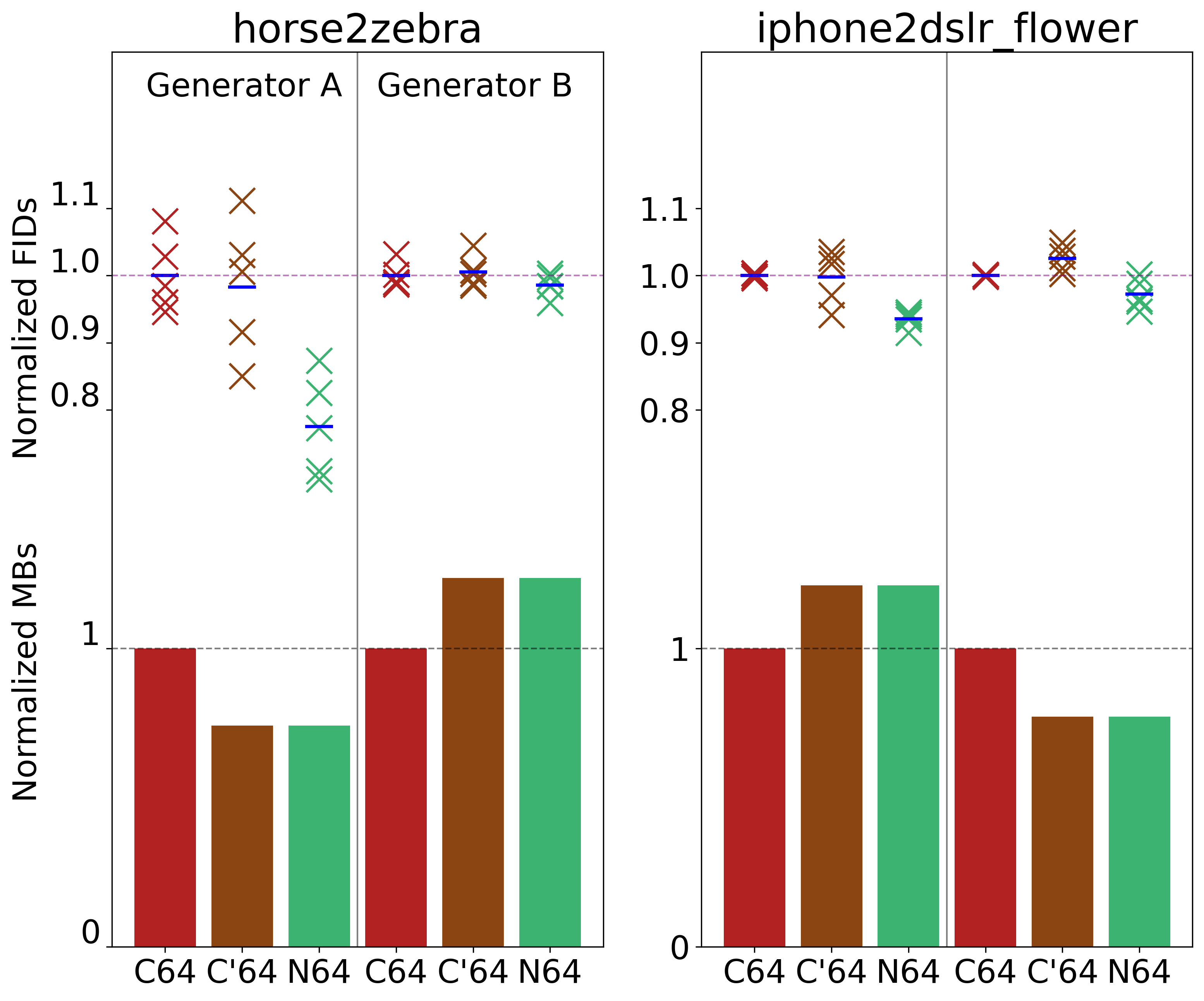}} 
    \caption{Performance of the original CycleGAN with $H=64$ ($C64$), its asymmetric variant by scaling the hidden dimension ($C'64$), and the architecture outcome of CycleGANAS ($N64$). Scaling the hidden dimension does not improve the performance, while the architecture search does. %\CJ{Hm.. the prime of $C'64$ is not much visible in the figure. Any suggestion? Shall we use $C64^*$ or $\widehat{C64}$?}\TG{enlarged the fonts and figure... }
    }
    \label{fig:CycleGAN_Asym}
    \vspace{-0.5cm}
    \end{center}
\end{figure}

% 1) performance of naively balanced CycleGAN 2) vs CUT, ...?
\subsection{Architecture search vs hidden dimension scaling}
The two generators of the original CycleGAN have the same architecture. In contrast, the architectures searched by CycleGANAS may have asymmetric structure that naturally comes from the dataset when it selects appropriate operations in $\mathcal{S}^e, \mathcal{S}^r, \mathcal{S}^d$ for each of the two generators. One may argue that the crucial factor to the performance is the asymmetric model size of the two generators, rather than the selection of their operations. We show that this is not the case, and the operation selection (i.e., architecture search) is of great importance.

For the datasets of horse2zebra and iphone2dslr, we consider the symmetric architectures of $C64$ (CycleGAN) and the asymmetric architectures $N64$ searched by CycleGANAS. Note that their total model sizes are the same. Then we build another asymmetric architectures $C'64$ by scaling the hidden dimension of $C64$ generators. To elaborate, we scale the hidden dimensions of two generators of $C64$ such that their model sizes equal to those of $N64$'s two generators, respectively. As a result, each generator of $C'64$ has the same operations as the $C64$ counterpart and the same model size as the $N64$ counterpart.

Fig.~\ref{fig:CycleGAN_Asym} shows the model sizes and the FIDs of $5$ evaluations, normalized w.r.t. the generator model size and mean FIDs of $C64$. It is confirmed that the generators of $C'64$ have the same model sizes as those of $N64$. We can observe that the mean FIDs ($\mathcal{A}$-to-$\mathcal{B}$, $\mathcal{B}$-to-$\mathcal{A}$) of $C'64$ are ($45.19, 132.87$) for horse2zebra and ($84.03, 93.41$) for iphone2dslr-flower, which are similar to $C64$'s ($45.78, 133.42$) and ($84.12, 91.70$), respectively. The searched asymmetric architectures of $N64$ achieve the lowest FIDs of ($38.05, 131.44$) and ($80.07, 89.75$), respectively. This suggests that, when dealing with data imbalance, simply scaling the hidden dimension might not yield substantial benefits; instead, identifying suitable operations becomes of paramount importance.

\section{Conclusion}
% CycleGANAS
We develop a NAS framework for CycleGAN that carries out unpaired image-to-image translation task. Compared to NAS for GANs, NAS for CycleGAN is more challenging due to the limited samples from dataset, multiple neural networks and their involvement in learning, and data imbalance.

% cells, single-level optimization
We design a framework, called CycleGANAS, that can search CycleGAN architectures of two generators and two discriminators, simultaneously. For flexible and practical search, we build architectures by stacking many simple ResNet-based cells, take the approach of differentiable search through super-networks, and apply the hidden dimension reduction. 
We further reduce the computational complexity and stabilize the search with the single-level optimization, enabling CycleGANAS to effectively explore a vast search space of size $8.1\times{10}^{36}$.

% performances, insights
Our experiments demonstrate that CycleGANAS effectively discovers high-performance architectures that either match or surpass the performance of the original CycleGAN. Furthermore, CycleGANAS successfully addresses the data imbalance by individually searching for distinct generator architectures for each translation direction, thereby regulating the model ratio.
{
    \small
    \bibliographystyle{ieeenat_fullname}
    \bibliography{main}
}


\end{document}